\documentclass{article}
\usepackage{iclr2027_conference,times}

\usepackage{amsmath,amsfonts,bm}

\def\eqref#1{equation~\ref{#1}}

\def\1{\bm{1}}

\DeclareMathAlphabet{\mathsfit}{\encodingdefault}{\sfdefault}{m}{sl}
\SetMathAlphabet{\mathsfit}{bold}{\encodingdefault}{\sfdefault}{bx}{n}

\usepackage{stmaryrd}
\usepackage{amssymb}
\usepackage{booktabs}
\usepackage{graphicx}
\usepackage{wrapfig}
\usepackage{microtype}
\usepackage{multirow}
\usepackage{xcolor}
\usepackage{tikz}
\usepackage{url}
\usepackage{hyperref}
\usepackage{xspace}

\usetikzlibrary{arrows.meta,positioning}

\definecolor{compilerblue}{RGB}{37,99,235}
\definecolor{compilerorange}{RGB}{234,88,12}

\newcommand{\tcenv}{\textsc{Triton Compiler Environment}\xspace}
\newcommand{\tcenvshort}{\textsc{TCEnv}\xspace}
\newcommand{\aicompiler}{\textsc{Triton AI Compiler}\xspace}
\newcommand{\aicompilershort}{\textsc{TAIC}\xspace}
\newcommand{\ailowering}{AI lowering\xspace}
\newcommand{\Ailowering}{AI lowering\xspace}

\newcommand{\featureused}{\textcolor{green!50!black}{\ensuremath{\checkmark}}}

\title{AI as a Compiler: Compiling Triton kernels without the Triton compiler}

\author{Francois Costa\thanks{Correspondence to: \texttt{francois.costa@stanford.edu}} \\
Stanford University
\And
Charly Castes \\
EPFL
\And
Thomas Bourgeat \\
EPFL
\And
Azalia Mirhoseini \\
Stanford University}

\iclrfinalcopy
\begin{document}

\maketitle

\lhead{}
\renewcommand{\headrulewidth}{0pt}

\begin{abstract}
Compiler backends are expensive to build and maintain as programming models, workloads, and accelerators evolve. We investigate whether large language models can replace the conventional optimizing and lowering pipeline, a process that we call \emph{\ailowering{}}.
We study \ailowering{} from Triton to NVIDIA PTX: an LLM agent translates Triton kernels directly into PTX.
We build an environment that evaluates candidate PTX, and an agentic harness in which an LLM translates Triton kernels into PTX. Across twelve common kernels on Ada, Hopper, and Blackwell GPUs and ten kernels from recent ML papers, \emph{\ailowering{}} achieves \(0.83\times\)-\(3.34\times\) the performance of autotuned Triton.
The largest gains come from transformations that Triton's lowering pipeline does not perform, such as decoding packed binary weights directly into Tensor Core operands (3.34$\times$ on BitDelta), assigning each thread a complete softmax row in tensor memory (1.37$\times$ on FlashAttention), and reusing overlapping convolution windows (up to 2.23$\times$). 
These results rely on a robust evaluation harness with comprehensive verification support. We build on Volta, an existing PTX verifier, and substantially extend it to support modern GPU architectures by introducing support for Blackwell's \texttt{tcgen05} Tensor Core interface. This requires modeling three architectural features: managed tensor memory, descriptor-based operand layouts, and asynchronous execution coordinated through commits, waits, memory barriers, and proxy fences. We discuss the challenges involved in formalizing them, as well as the current limitations. Our results suggest an emerging future in which AI compilers replace custom-written intermediate representations and checkers, reducing the time and engineering effort required to bring up software for new general-purpose and custom chips.
\end{abstract}

\section{Introduction}
\label{sec:introduction}

High-performance GPU kernels require substantial hardware-specific expertise, making them costly to develop, optimize, and maintain as workloads and architectures evolve. Domain-specific languages such as Triton, TileLang, CuTe DSL, ThunderKittens, HipKittens, and ParallelKittens reduce the burden on kernel programmers by exposing tile-oriented or hardware-aware abstractions \citep{tillet2019triton,wang2025tilelang,cute2025dsl,spector2025thunderkittens,hu2025hipkittens,sul2025parallelkittens}. However, they largely shift complexity into the compiler: supporting new abstractions and hardware features still requires implementing lowering, scheduling, legality checks, and optimizations. Traditional compiler backends further amplify this burden by relying on multiple intermediate representations (IRs), whose semantics, transformations, and lowering paths must be continuously extended and maintained. Systems such as MLIR aim to reduce duplicated infrastructure, yet large production compilers such as LLVM and XLA still illustrate how difficult it is to maintain extensive optimization and code-generation pipelines without introducing correctness or performance regressions \citep{lattner2021mlir,mannarswamy2022combine,openxla2024emitters}.

This problem is becoming more severe as accelerator architectures diversify and evolve rapidly. GPUs, TPUs, Trainium, wafer-scale systems, LPUs, in-memory-compute processors, and custom AI ASICs expose increasingly different execution, memory, and communication models, while even successive generations of the same accelerator family can introduce capabilities that existing IRs cannot naturally express. Consequently, new systems increasingly introduce specialized abstractions and compilation paths, such as Tawa's asynchronous references for warp specialization or ThunderKittens' hardware-specific tile and warp abstractions \citep{chen2026tawa,spector2025thunderkittens}. This rapid pace of hardware innovation makes conventional IR-centric compiler development increasingly expensive and difficult to keep aligned with state-of-the-art performance. When existing compiler abstractions cannot expose new hardware efficiently, performance-critical implementations bypass the IR using target-specific intrinsics or assembly, as DeepGEMM does with Hopper-specific PTX instructions \citep{deepseek2025deepgemm,nvidia2026ptx}.

\begin{figure*}[t]
  \centering
  \includegraphics[width=\textwidth]{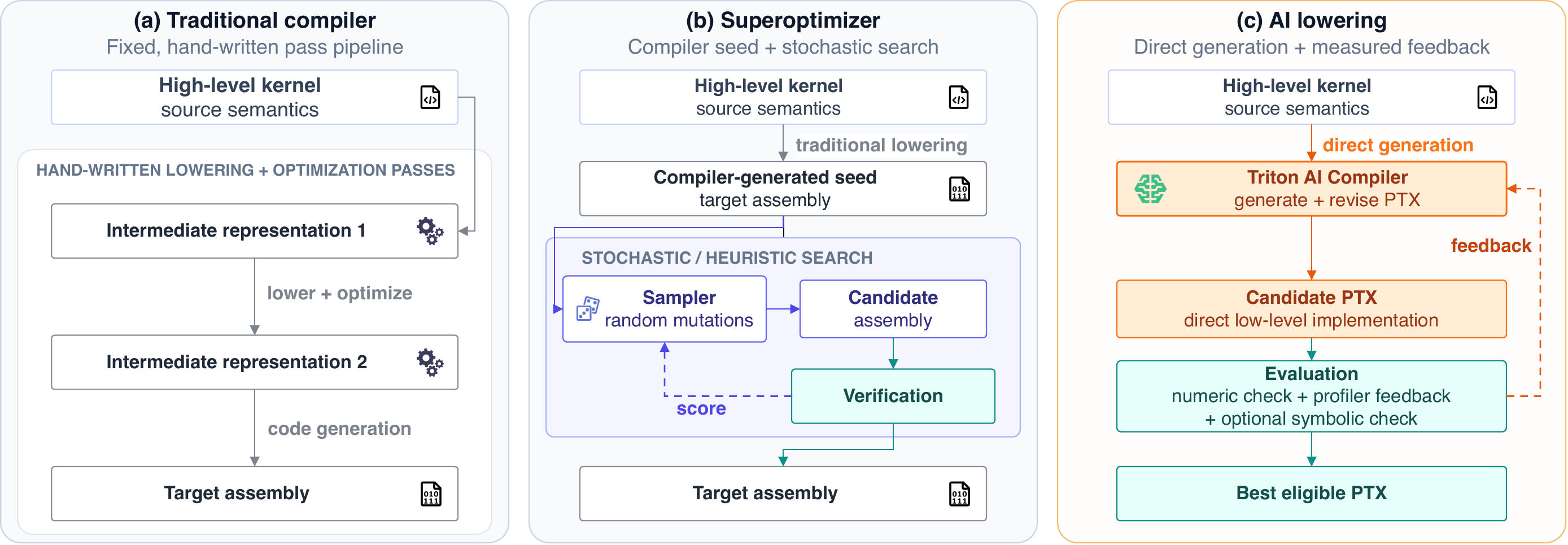}
  \caption{Traditional compilation, superoptimization, and \ailowering{}.
  (a) A traditional compiler applies hand-written lowering and optimization
  passes across intermediate representations. (b) A superoptimizer uses
  conventionally lowered target code as a seed, then iteratively mutates and
  scores candidate programs. (c) \Ailowering{} bypasses the conventional
  optimizing/lowering backend: an LLM generates PTX
  directly and revises it using numerical and performance feedback, and optionally
  symbolic verification. In the search-based approaches, acceptance is guarded by
  numerical or formal checks.}
  \label{fig:compilation-paradigms}
\end{figure*}

We show that, for critical GPU kernels, \textbf{bypassing the conventional lowering pipeline is not only feasible but beneficial}.
We introduce the \aicompiler{} (\aicompilershort{}), a direct Triton-to-PTX compiler that realizes \ailowering{}: an LLM synthesizes candidate PTX implementations directly from the Triton source. This contrasts with conventional compilers, which progressively lower high-level code through multiple hand-designed intermediate representations.
We observe that \ailowering{} effectively optimizes PTX code in non-trivial ways that are inaccessible to the baseline Triton compiler, such as combining Tensor Core instructions with warp-level shuffles, restructuring data movement and computation directly at the PTX level, and selecting target-specific instructions that are not exposed by Triton's evaluated lowering pipeline.
Across Ada, Hopper, and Blackwell GPUs, \ailowering{} achieves between 0.83$\times$ and 3.34$\times$ the performance of optimized Triton-generated PTX.
On B200, it also outperforms widely used optimized Triton kernels, including by 1.37$\times$ on FlashAttention and 1.10$\times$ on a Mamba-2 primitive.

While generation is inherently probabilistic, correctness should not be, yet existing verification tools cover only a fragment of modern GPU execution. Volta~\citep{volta}, the state of the art for PTX verification, symbolically executes every thread in a CTA and checks races, deadlocks, and output expressions, but primarily supports registers, shared memory, warp primitives, bulk-synchronous barriers, pre-Hopper instructions, and input-independent control flow. Modern kernels exceed this fragment, while \Ailowering{} widens the gap further, using more than 100 PTX instruction variants that do not appear in the original output of the Triton compiler. Restricting \ailowering{} to the verifier-supported subset erases its performance advantage, so we view this gap between rapidly evolving GPU ISAs and formally specified semantics as the new central challenge for AI kernel generation. In this spirit, we build an extension for Volta supporting Hopper and Blackwell. This includes integrating Blackwell's \texttt{tcgen05} interface, which adds tensor memory, descriptor-based layouts, asynchronous collective issue, \texttt{mbarrier} transaction accounting, and proxy fences~\citep{nvidia2026ptx}, whose symbolic effects and ordering guarantees both need modeling. Finally, we discuss the challenges that remain in proving the soundness of the verifier.

Our contributions are as follows:

\begin{itemize}
    \item We introduce the \aicompiler{} (\aicompilershort{}) and the \tcenv{} (\tcenvshort{}),
    combining LLM-based PTX generation with compilation feedback, correctness
    testing, benchmarking, profiling, and iterative optimization, while
    preserving the kernel ABI\footnote{Application Binary Interface.}, launch configuration, compile-time constants,
    and target-GPU contract.
    \item We show that generated PTX achieves $0.83\times$-$2.23\times$ Triton performance
    on our twelve-kernel suite across Ada, Hopper, and Blackwell,
    uses architecture-specific primitives, and outperforms Triton on recent
    kernels, including BitDelta ($3.34\times$), FlashAttention ($1.37\times$),
    and Mamba-2 ($1.10\times$) on B200.
    \item Collectively, our results suggest a future in which AI compilers could replace custom-written intermediate representations with end-to-end, direct translation from high-level source to low-level code, lowering the complexity and time required to bring up software for new chips and architectures. We demonstrate that realizing this vision requires substantially stronger verification and identify the key remaining challenges through our extension of Volta to Blackwell.
\end{itemize}

\section{Related Work}
\label{sec:related}

\paragraph{Tensor compilers, GPU DSLs, and kernel libraries.}
TVM and Ansor separate tensor computations from schedules and search over
scheduling choices, while Mirage superoptimizes across kernel, thread-block,
and thread levels~\citep{chen2018tvm,zheng2020ansor,wu2025mirage}. Triton and
Halide instead compile higher-level programming models, and CUTLASS and
ThunderKittens provide expert-designed implementations and
abstractions~\citep{tillet2019triton,ragankelley2013halide,nvidia2026cutlass,spector2025thunderkittens}.
Rather than searching a hand-engineered schedule space or selecting a library
implementation, \ailowering{} synthesizes hardware-specific PTX directly
from a fixed Triton kernel.

\paragraph{LLM-generated GPU kernels.}
KernelBench and TritonBench evaluate CUDA or Triton generation from
PyTorch-level tasks~\citep{ouyang2025kernelbench,li2025tritonbench}. AI CUDA
Engineer adds evolutionary search~\citep{lange2025aicudaengineer}, CUDA-L1 uses
contrastive reinforcement learning, and Dr. Kernel studies multi-turn
reinforcement learning with methods designed to limit reward hacking and biased
gradients~\citep{li2026cudal1,liu2026drkernel}. TritonGym
provides a benchmark and tool-orchestration framework for agentic Triton
generation~\citep{guan2026tritongym}, whereas PTXBench generates CUDA augmented with
prescribed architecture-specific PTX instructions~\citep{zhang2026ptxbench}.
Our task instead requires PTX-only output for a fixed, self-contained Triton
program and leaves instruction selection unconstrained within the compilation
contract.

\paragraph{Superoptimization and learned compilation.}
STOKE searches loop-free x86 binaries, while SuperCoder uses LLMs and
reinforcement learning to optimize larger x86-64 assembly
programs~\citep{schkufza2013stoke,wei2026supercoder}. Other work builds on explicit IRs:
MLIR provides multi-level compiler infrastructure, CompilerGym exposes compiler
decisions to agents, LLM Compiler learns LLVM IR and assembly representations,
and CAKE co-designs an agent-facing GPU IR and compiler
harness~\citep{lattner2021mlir,cummins2022compilergym,cummins2024llmcompiler,ye2026cake}.
In contrast, \ailowering{} starts from Triton rather than target assembly
and bypasses the conventional IR stack while jointly performing instruction
selection, allocation, and scheduling in PTX.

\paragraph{Correctness of generated code.}
KernelBench and related systems rely on finite test
suites~\citep{ouyang2025kernelbench,zhang2026kernelbenchverified,wu2025mirage}.
Volta symbolically executes PTX and proves
equivalence~\citep{volta}, but excludes the PTX instructions introduced by
Hopper and Blackwell.

\section{An Alternative Compilation Route}
\label{sec:problem}

We choose Triton-to-PTX to explore \ailowering{} because it provides a clean kernel-level compilation boundary. Starting from PyTorch would introduce graph-level optimizations such as fusion, decomposition, and library dispatch, making kernel-level gains harder to isolate. CUDA, conversely, already exposes many low-level choices explicitly. Triton offers a useful middle ground: it specifies the computation and kernel interface while leaving hardware-specific decisions such as instruction selection, thread mapping, data movement, synchronization, and scheduling to the compiler.

\subsection{Problem Formulation}

A Triton kernel is compiled under a fixed contract that determines both its specialization and execution. For each compilation instance, specialization corresponds to fixed values such as tensor dimensions, strides, and \texttt{constexpr} parameters (\textit{the compile-time values}). The runtime launch interface (\textit{the runtime interface}) specifies the grid dimensions, number of threads, and argument ABI. Finally, the generated code must target a specific GPU architecture (\textit{the underlying hardware}). \Ailowering{} must preserve this contract while replacing the conventional optimizing and lowering backend with model-generated PTX.

Formally, let $K$ be a Triton kernel and let $(c,I,h)$ denote its execution contract, where $c$ denotes the fixed specialization parameters (\textit{the compile-time values}), $I$ denotes the launch configuration and argument ABI (\textit{the runtime launch interface}), and $h$ denotes the target GPU architecture (\textit{the underlying hardware}). The contract induces a reference input-output semantics
$\llbracket K \rrbracket_{c,I,h}$ over the legal runtime inputs $\mathcal{D}$; it does not designate a unique PTX program. A candidate PTX program $P$ is correct when it exposes the required entry point and argument ABI and, under the same contract $(c,I,h)$, satisfies
\begin{equation}
    \llbracket P \rrbracket_{c,I,h}(x)
    \approx_{\epsilon}
    \llbracket K \rrbracket_{c,I,h}(x),
    \qquad \forall x \in \mathcal{D}.
    \label{eq:equivalence}
\end{equation}

Here $\approx_{\epsilon}$ denotes numerical equivalence within predefined absolute and relative tolerances; it allows small floating-point rounding differences without requiring bitwise identity. When formal verification is available, we additionally check exact equality after abstracting floating-point operations as real arithmetic. This check captures the intended arithmetic semantics rather than a particular sequence of compiler-generated floating-point operations.

\Ailowering{} is an iterative synthesis process rather than a deterministic
compilation procedure. At iteration $t$, the harness constructs a prompt
$q_t=\operatorname{Prompt}(K,c,I,h,\mathcal{F}_{<t})$ using the compilation contract and,
when available, feedback from earlier candidates, then queries the language model:
\begin{equation}
  P_t \sim p_{\theta}(\,\cdot\mid q_t\,),
  \label{eq:lm-query}
\end{equation}
where $p_{\theta}$ is the model's conditional distribution.

Let $\mathrm{PTX}$ denote the space of PTX programs. The compilation contract
defines the set of all valid lowerings of $K$ for the fixed compile-time values
$c$, runtime interface $I$, and target $h$ as
\begin{equation}
  \mathcal{V}(K,c,I,h)
  = \left\{
      P \in \mathrm{PTX}
      \;\middle|\;
      \substack{
        \operatorname{Compiles}(P,h) \\
        {}\land\ \operatorname{Safe}(P) \\
        {}\land\ \operatorname{ABI}(P)=\operatorname{ABI}(K,I) \\
        {}\land\ \forall x\in\mathcal{D}:\
        \llbracket P \rrbracket_{c,I,h}(x)\approx_{\epsilon}\llbracket K \rrbracket_{c,I,h}(x)
      }
    \right\},
  \label{eq:valid-lowerings}
\end{equation}
where $\operatorname{Compiles}(P,h)$ requires $P$ to be a legal PTX program
for target $h$, while $\operatorname{Safe}(P)$ requires memory safety and
race freedom. Let $\widehat{L}(P;h)$ be the latency estimated by the fixed
benchmarking protocol. The objective is
$\min_{P\in\mathcal{V}(K,c,I,h)}\widehat{L}(P;h)$.

The remainder of the paper makes this formulation operational. Section~\ref{sec:tcenv} introduces \tcenvshort{}, our environment for evaluating PTX. Section~\ref{sec:llm-compilation-process} describes \aicompilershort{}, the harness that performs the iterative compilation process and NCU-guided optimization. Section~\ref{sec:formal-verification} discusses verification and the limits of what can currently be verified soundly.

\subsection{The \tcenv{}: An environment for \ailowering{}}
\label{sec:tcenv}

\begin{figure*}[t]
  \centering
  \includegraphics[width=\textwidth]{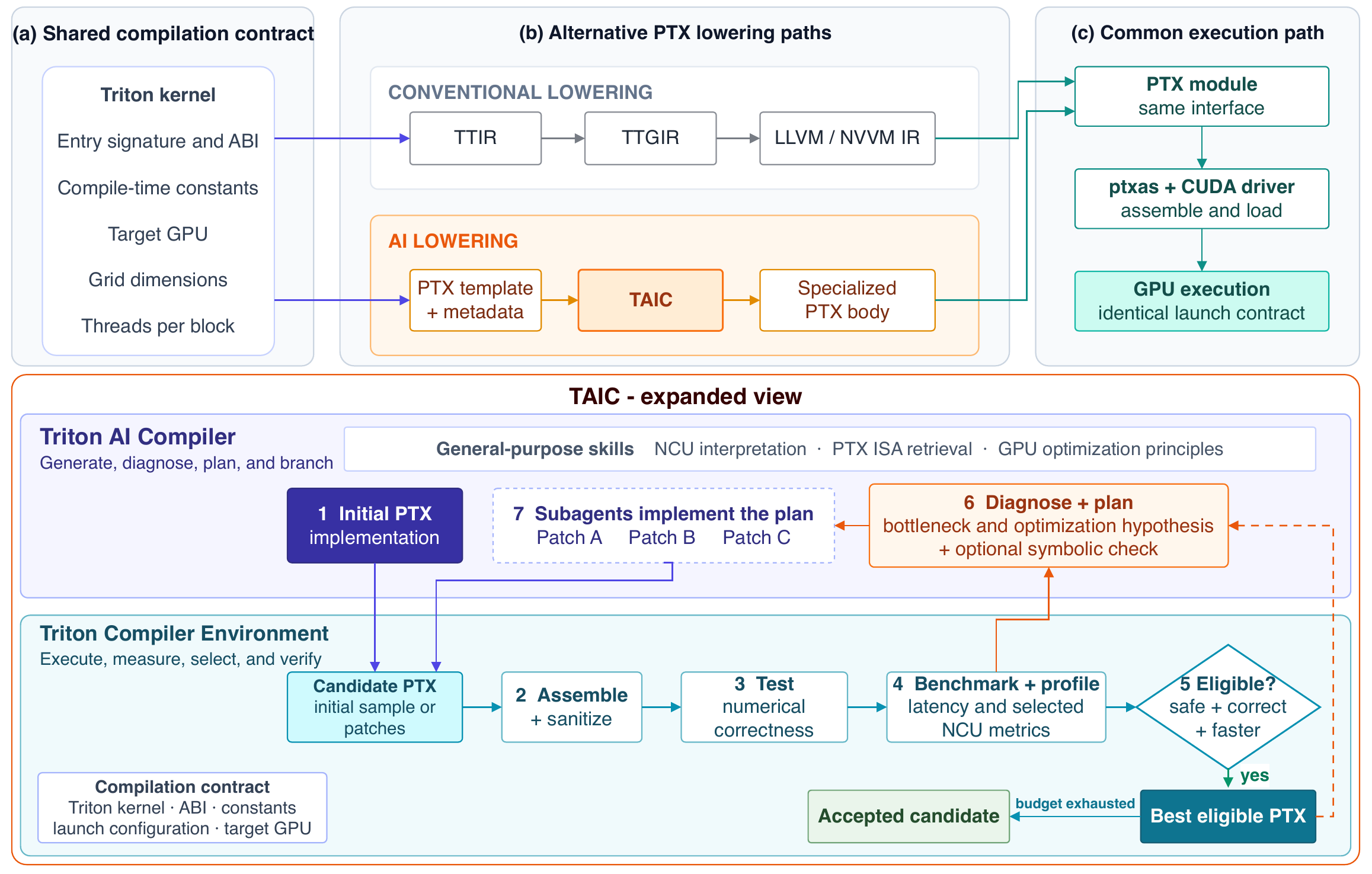}
  \caption{System overview and detailed AI-compilation loop. Top:
  conventional compilation and \ailowering{} start from the same compilation contract,
  produce an interface-compatible PTX module, and share the NVIDIA execution
  path. The orange block expands \aicompilershort{} into the detailed loop below,
    where \aicompilershort{} proposes and revises PTX while \tcenvshort{} assembles,
  tests, benchmarks, profiles, and selects candidates.}
  \label{fig:compilation-paths}
\end{figure*}

We build the \tcenv{} (\tcenvshort{}), a controlled environment that allows \aicompilershort{} to focus exclusively on \ailowering{}. It abstracts away configuration details that are orthogonal to lowering, such as compile-time constants, grid dimensions, threads per block, and the target GPU architecture: these are provided to Triton as fixed inputs during compilation, so they need not be inferred as part of \ailowering{}. Concretely, \tcenvshort{} exposes a dictionary containing the Triton kernel, its hyperparameters, and execution metadata, together with a PTX template that the launcher can invoke without further configuration, with the function signature and arguments preserved in the PTX interface. It also provides the infrastructure required to assemble, launch, validate, and profile generated PTX. \aicompilershort{} therefore receives a high-level kernel representation and produces a semantically equivalent low-level one, as defined in the previous section; Figure~\ref{fig:compilation-paths} illustrates this path alongside the standard compilation passes.

An artificially weak baseline can make performance gains appear much larger than they are, effectively a form of reward hacking. This matters because semantically equivalent implementations can differ significantly in performance, as the fused softmax example in the Triton tutorial illustrates~\citep{tritonFusedSoftmaxTutorial}. We therefore construct each Triton reference following the optimization best practices of the official tutorials~\citep{tritonTutorials}, and, before \tcenvshort{} exposes a reference kernel, autotune it on the target GPU and use the resulting configuration throughout the evaluation.

\tcenvshort{} combines randomized differential testing with optional symbolic verification (Section~\ref{sec:formal-verification}) to obtain correctness signals. The randomized tester generates inputs that stress overflow, underflow, cancellation, and precision loss, and rejects kernels whose outputs deviate from the reference beyond the prescribed tolerance. The symbolic verifier instead checks that the generated program implements the intended computation. 

\subsection{The \aicompiler{}: An agent performing \ailowering{}}
\label{sec:llm-compilation-process}

We build the \aicompiler{} (\aicompilershort{}), a harness that operates through
\tcenvshort{} (Section~\ref{sec:tcenv}) to realize a direct Triton-to-PTX
compilation path that bypasses Triton's intermediate compiler representations.
\aicompilershort{} is an LLM augmented by three general-purpose skills:
interpreting Nsight Compute (NCU) metrics, retrieving PTX ISA documentation, and
applying high-level GPU optimization principles. It performs generation and
optimization, whereas \tcenvshort{} provides the execution and verification
infrastructure, a division of responsibilities made explicit in
Figure~\ref{fig:compilation-paths}.

\aicompilershort{} first produces an initial PTX implementation, which
\tcenvshort{} assembles, sanitizes, checks for numerical correctness, and
benchmarks. Inspired by verifier-backed inference-time
scaling~\citep{brown2024largelanguagemonkeys, ouyang2025fastkernels}, we then refine
it: given the current PTX together with assembly and sanitizer outcomes, benchmark
results, and selected profiler observations, \aicompilershort{} diagnoses the
dominant bottleneck and formulates an optimization plan, ranging from local
instruction-level edits to global changes in thread mapping, data movement, or
synchronization. Subagents realize this plan independently as candidate patches,
which \tcenvshort{} evaluates under the same contract; only candidates that
assemble, pass the safety and correctness gates, and improve the measured objective
are eligible to seed the next round. The best eligible implementation and its
measurement feedback return to \aicompilershort{}, which repeats this
diagnose-plan-branch-evaluate cycle until the search budget is exhausted.

\section{Evaluation}
\label{sec:methodology}

We use GPT-6 Astra as the compiler LLM. Each kernel is evaluated on three NVIDIA GPU generations: an L40S (Ada Lovelace, \texttt{sm\_89}), an H100 (Hopper, \texttt{sm\_90a}), and a B200 (Blackwell, \texttt{sm\_100a}). Figure~\ref{fig:architecture-taxonomy} summarizes the architecture-specific features present on these targets. Experiments run sequentially on otherwise idle GPUs. Generated code targets PTX ISA 8.7 with a 64-bit address space and is assembled for the corresponding architecture.

\begin{figure*}[h]
\centering
\resizebox{\textwidth}{!}{%
\renewcommand{\arraystretch}{1.1}
\begin{tabular}{@{}lcccccc@{}}
\toprule
& \textbf{Ada} & \textbf{used?}
& \textbf{Hopper} & \textbf{used?}
& \textbf{Blackwell} & \textbf{used?} \\
\midrule
\textbf{MMA} & \texttt{mma.sync} + \texttt{ldmatrix} & \featureused
& \texttt{wgmma} + TMA loads & \featureused
& \texttt{tcgen05} + TMEM accum. & \featureused \\
\textbf{SMEM} & XOR bank swizzle & \featureused
& 128-B XOR swizzle & \featureused
& 64-B offset swizzle & \featureused \\
\textbf{Copies} & \texttt{cp.async} pipeline & \featureused
& TMA + \texttt{mbarrier} & \featureused
& TMA 2-D + phase parity & \featureused \\
\bottomrule
\end{tabular}
}
\caption{Architecture-specific features and their observed use in
LLM-generated PTX. A check mark denotes that the generated kernels use the
feature on the corresponding microarchitecture. The Hopper and Blackwell
entries (\texttt{wgmma}, TMA, \texttt{mbarrier}, \texttt{tcgen05}, TMEM) fall
outside Volta's originally validated semantics; verifying kernels that use
them relies on the emulator extension of
Section~\ref{sec:formal-verification}.}
\label{fig:architecture-taxonomy}
\end{figure*}

Our baseline uses Triton 3.6.0 and draws its kernels from TritonBench and
OpenAI's official Triton tutorials~\citep{li2025tritonbench,tritonTutorials}.
To ensure a competitive baseline, we further optimize these kernels where
applicable, including rewriting kernels when faster implementations can be
obtained and removing boundary masks when the evaluated dimensions guarantee
in-bounds accesses. For each kernel and GPU, we report the fastest correct
configuration found through autotuning. To keep the comparison representative
of standard Triton programming, the baseline kernels do not use experimental
Triton features or manually written inline assembly.

We use GPT-6 Astra throughout our main evaluation because our model comparison shows that competitive \ailowering{} emerges only with the latest frontier models. Under the same protocol and inference budget on FP16 GEMM, GPT-5.4 reaches $0.73\times$ Triton performance, GPT-5.5 and GPT-5.6 Terra reach $0.87\times$, GPT-5.6 Luna fails to produce an accepted kernel, and GPT-5.6 Sol reaches $0.97\times$, while GPT-6 Astra is the first model to reach parity with Triton at $1.00\times$. Since Triton already provides a highly optimized baseline for this workload, we select GPT-6 Astra as the representative model for our experiments: it is the first model in our comparison for which direct \ailowering{} is consistently competitive with conventional compilation.

We evaluate kernels spanning four performance-critical LLM workloads: attention, GEMMs, activation functions, and normalization or regularization operations. These stress complementary aspects of GPU optimization, including Tensor Core utilization, reductions and synchronization, memory access, and instruction throughput. We use fixed problem sizes representative of dimensions commonly encountered in LLM workloads.

\subsection{Architectural Coverage and Source Semantics Understanding}
\label{sec:architecture-and-semantics}

\paragraph{Architectural coverage.}
Figure~\ref{fig:architecture-taxonomy} shows that the generated PTX uses each
generation's modern mechanisms rather than a lowest-common-denominator
instruction set: \texttt{mma.sync} with \texttt{ldmatrix} and
\texttt{cp.async} pipelining on Ada, \texttt{wgmma} with TMA loads and
\texttt{mbarrier} synchronization on Hopper, and \texttt{tcgen05} with
tensor-memory accumulators and two-dimensional TMA transfers on Blackwell;
shared-memory layouts likewise progress from XOR bank swizzling to 128-byte
XOR and 64-byte offset swizzles. This coverage is also what makes verification
difficult: the Hopper and Blackwell mechanisms fall outside the semantics
originally validated by Volta, and reaching them requires the emulator
extension of Section~\ref{sec:formal-verification}.

\paragraph{Recovering semantics from source.}
Because many of our Triton kernels may have appeared in the model's training data, we evaluate whether \aicompilershort{} actually recovers source semantics rather than relying on memorized implementations or familiar high-level cues. We first test randomly generated kernels of roughly 150 lines that implement no recognizable algorithm, operate on integers to avoid floating-point discrepancies, and exercise a broad range of Triton features, including integer and bitwise arithmetic, modulo and hashing-like computations, reductions such as \texttt{tl.sum} and \texttt{tl.max}, \texttt{tl.static\_range} loops, \texttt{tl.where}, and hints such as \texttt{tl.multiple\_of} and \texttt{tl.max\_contiguous}; \aicompilershort{} produces correct PTX for every one. We then introduce deliberately misleading cues: a matrix multiplication whose accumulation is changed to $c[i,j] \mathrel{-}= a[i,k]\cdot b[k,j]$, a FlashAttention variant with \texttt{qk\_scale} $=$ \texttt{HEAD\_DIM}$^{-0.5} + 1.4426950408889634$, and a kernel named \texttt{ReLU} whose comments request GELU. In all three cases, the model follows the actual implementation rather than the name, comments, or familiar algorithmic structure, producing semantically equivalent PTX. Finally, when the Triton source requests IEEE-754 semantics, the generated PTX avoids Tensor Core operations, further indicating that the model captures semantic constraints expressed in the source rather than merely reproducing common optimized patterns.

\begin{figure*}[t]
\centering
\includegraphics[width=\textwidth]{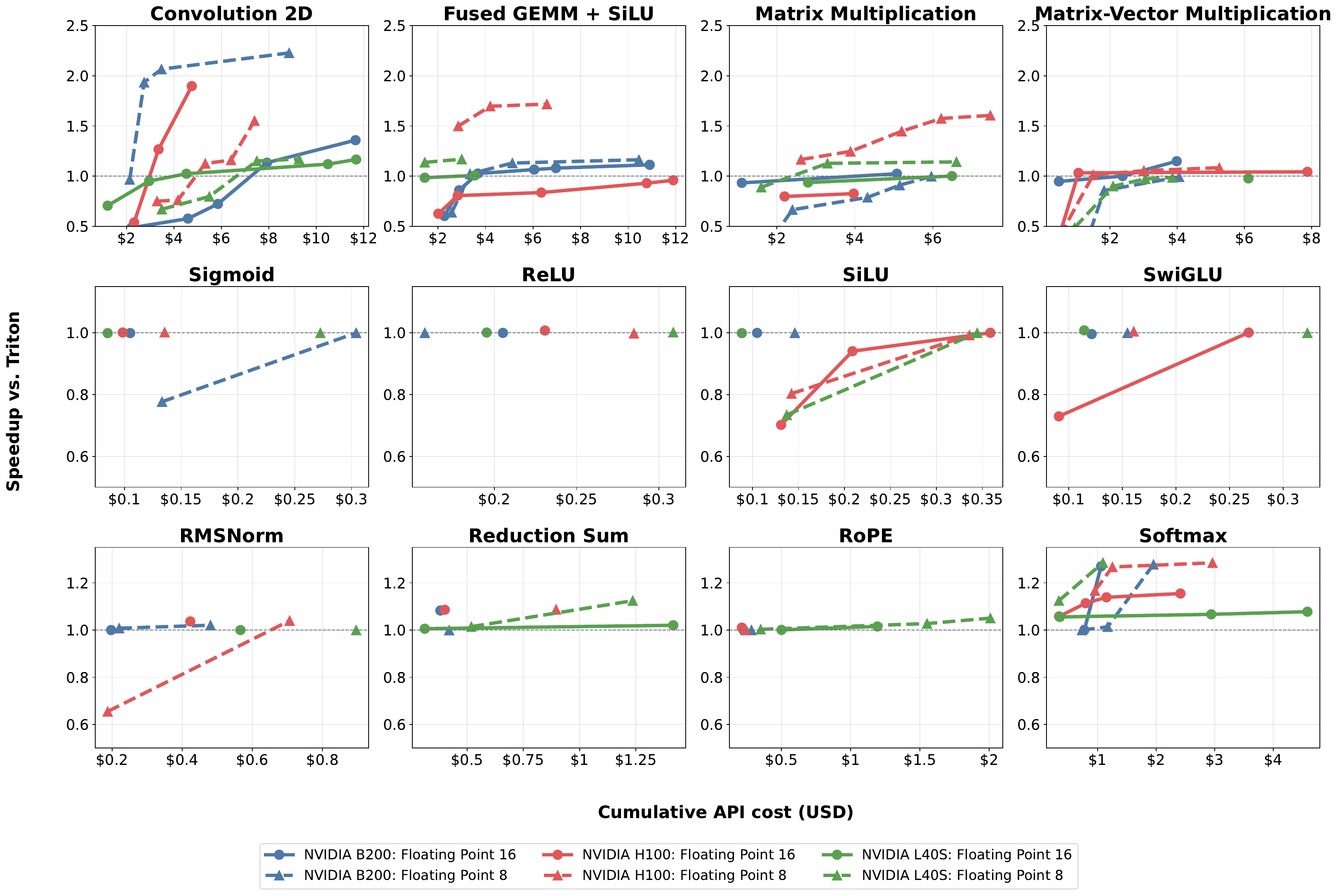}
\caption{Per-kernel speedup/cost Pareto frontiers relative to Triton for FP16
and FP8 on three GPU architectures. Each point is a validated candidate that
improves the best speedup reached at its cumulative API cost; the dashed line
marks parity with the corresponding autotuned Triton baseline. Axes are scaled
independently across panels.}
\label{fig:kernel-results}
\end{figure*}

\subsection{Results on Common Kernels}

Figure~\ref{fig:kernel-results} shows that the benefit of \ailowering{}
depends strongly on the operator, precision, and target GPU. Sigmoid, ReLU,
SiLU, and SwiGLU finish within approximately $1\%$ of Triton, and RMSNorm improves by at
most $4\%$. These panels therefore primarily demonstrate recovery of baseline
performance; their narrow vertical scales should not be read as large gains.
RoPE improves by at most $5\%$, and matrix-vector
multiplication ranges from $0.98\times$ to $1.15\times$.
Softmax improves by $1.08$-$1.29\times$, whereas convolution reaches
$1.17$-$2.23\times$. GEMM is more architecture- and precision-dependent:
FP16 reaches parity on B200 and L40S but only $0.83\times$ on H100, while
FP8 reaches $1.61\times$ on H100, $1.15\times$ on L40S, and approximately
parity on B200. The fused GEMM-SiLU kernel exhibits the same H100 contrast
($0.96\times$ for FP16 versus $1.72\times$ for FP8).
The cost axis also distinguishes easy recoveries from substantial searches:
elementwise candidates generally cost less than \$1.50, while convolution
and GEMM candidates require roughly \$5-\$12. The main grid reports a single
run per experiment; Section~\ref{sec:repeatability} separately repeats four
B200 workloads five times. To provide concrete examples of the optimizations
discovered by \aicompilershort{}, we briefly highlight three representative kernels.
For reduction sum, the speedup comes from replacing Triton's
generic reduction with architecture-specific Tensor Core or warp-reduction
algorithms. For convolution, it comes from reusing inputs across overlapping
windows and overlapping packed data movement with Tensor Core execution. For
H100 FP8 GEMM, it comes from using larger WGMMA tiles, fewer synchronization
points, and a cheaper transposing epilogue. See Appendix~\ref{app:optimization-examples}
for detailed comparisons with Triton.

\subsection{Repeatability under a Fixed Search Budget}
\label{sec:repeatability}

To measure run-to-run variation, we repeat four representative B200
experiments five times with independent model samples. We configure a maximum
API budget of \$15 per run for convolution, matrix multiplication, and
FlashAttention, and \$1 for the simpler elementwise SwiGLU kernel. The budget
gate is evaluated before issuing each model request; because a request is
atomic, its completion can carry the final recorded cost beyond the configured
gate. Table~\ref{tab:five-trial-speedups} reports the best validated candidate
from each run in chronological order.
We observe that the \aicompiler{} matches or outperforms the Triton compiler, although with moderate variance for kernels with significant gains over the baseline.

\begin{table*}[h]
\centering
\small
\setlength{\tabcolsep}{5pt}
\resizebox{\textwidth}{!}{%
\begin{tabular}{@{}lcrrrrrrr@{}}
\toprule
\textbf{Kernel} & \textbf{Budget} & \textbf{Trial 1}
& \textbf{Trial 2} & \textbf{Trial 3} & \textbf{Trial 4}
& \textbf{Trial 5} & \textbf{Mean $\pm$ SD} & \textbf{Best} \\
\midrule
Convolution (FP8) & \$15 & $2.081\times$ & $1.706\times$ & $2.229\times$ & $1.933\times$ & $2.070\times$ & $2.004 \pm 0.197$ & $2.229\times$ \\
Matrix multiplication (FP16) & \$15 & $1.002\times$ & $1.026\times$ & $1.032\times$ & $1.029\times$ & $1.024\times$ & $1.023 \pm 0.012$ & $1.032\times$ \\
SwiGLU (FP16) & \$1 & $1.000\times$ & $1.002\times$ & $1.000\times$ & $1.000\times$ & $1.000\times$ & $1.000 \pm 0.001$ & $1.002\times$ \\
FlashAttention forward (FP16) & \$15 & $1.261\times$ & $1.075\times$ & $1.214\times$ & $1.263\times$ & $1.366\times$ & $1.236 \pm 0.106$ & $1.366\times$ \\
\bottomrule
\end{tabular}
}
\caption{Repeatability on NVIDIA B200 across five independent
\aicompiler{} runs. Values are speedups over the same autotuned
Triton baseline. Budget is the configured per-run API-cost gate, checked
before each new model request; an in-flight request can finish above it.}
\label{tab:five-trial-speedups}
\end{table*}

\subsection{Results on Recently Published Kernels}

\begin{figure*}[t]
\centering
\includegraphics[width=\textwidth]{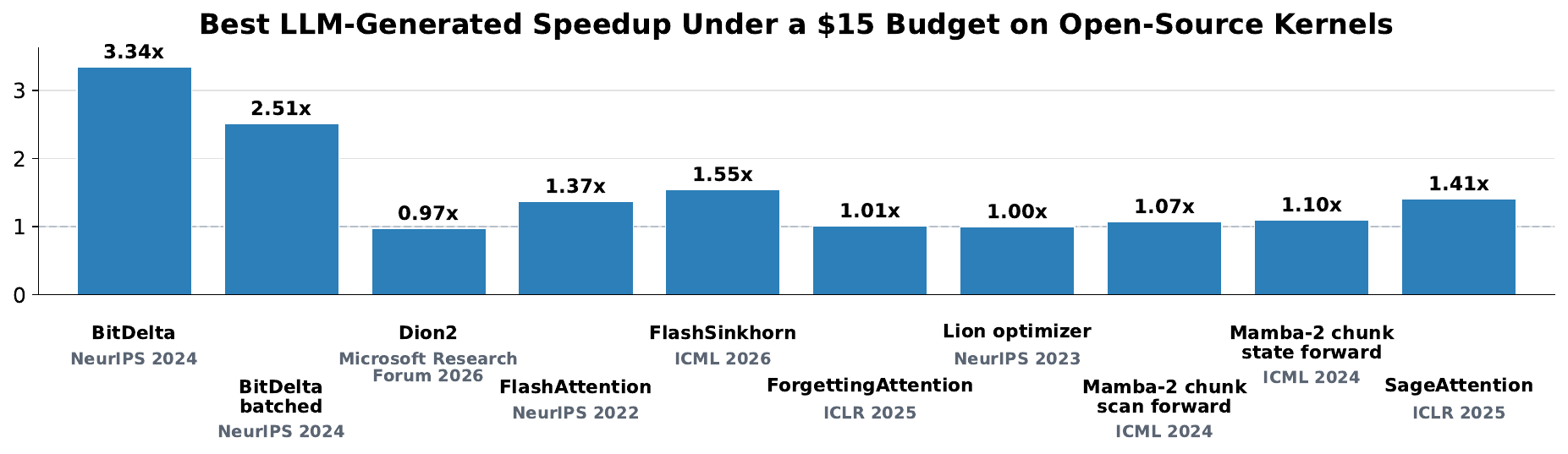}
\caption{Best numerically validated LLM-generated kernels on NVIDIA B200.
Each bar reports the best accepted candidate for one kernel; the dashed line
marks parity with Triton.}
\label{fig:conference-kernel-speedups}
\end{figure*}

Beyond the twelve-kernel suite of Section~\ref{sec:methodology}, we evaluate ten kernels from recent ML papers: four attention variants (FlashAttention~\citep{dao2022flashattention}, FlashSinkhorn~\citep{ye2026flashsinkhorn}, Forgetting Attention~\citep{lin2025forgetting}, and SageAttention~\citep{zhang2025sageattention}), two BitDelta GEMM variants~\citep{liu2024bitdelta} (standard and batched matrix multiplication), two memory-bound kernels from Lion~\citep{chen2023lion} and Dion~\citep{ahn2025dion}, and two Mamba-2 primitives~\citep{dao2024mamba2}. The Mamba-2 kernels differ substantially from attention, propagating compact recurrent state rather than forming pairwise token interactions. We use the authors' official open-source implementations without modifying their algorithms or optimizations, except for removing unnecessary masks, and autotune relevant parameters to obtain strong Triton baselines. Figure~\ref{fig:conference-kernel-speedups} summarizes the results.

Under a \$15 budget, \ailowering{} substantially outperforms Triton on several workloads, most notably the two BitDelta variants ($3.34\times$ and $2.51\times$), FlashSinkhorn ($1.55\times$), SageAttention ($1.41\times$), and FlashAttention ($1.37\times$). Memory-bound and recurrent workloads remain near parity, including Forgetting Attention ($1.01\times$), Lion ($1.00\times$), Dion ($0.97\times$), and Mamba-2 ($1.07\times$ and $1.10\times$). The largest gains therefore arise when generated PTX can reorganize computation and data movement. BitDelta decodes packed binary weights directly into Tensor Core operands; FlashAttention maps complete score rows to threads in the native tensor-memory layout, avoiding cross-lane reductions; and FlashSinkhorn performs independent warp-local log-sum-exp reductions while directly issuing FP16 MMA instructions. Appendix~\ref{app:optimization-examples} provides detailed comparisons.

\providecommand{\TODO}[1]{\textcolor{red}{\textbf{TODO: #1}}}

\subsection{Formal Verification in the Loop}
\label{sec:formal-verification}
When we activate the use of Volta~\citep{volta} in the compilation loop, a candidate is valid only
if it passes the numerical check \emph{and} Volta proves it equivalent to the
Triton reference.
When a kernel uses a pattern outside Volta's support,
\textsc{TCEnv} reports the unsupported feature to \aicompilershort{}, which then revises the candidate.
Table~\ref{tab:verified-vs-unverified-speedups}
shows that the resulting kernels are significantly less competitive: the
features that deliver performance on modern GPUs are not yet all supported by Volta.

To enable the use of modern GPU features, we extend Volta with support for the unsupported features that \aicompilershort{} attempts to use when left to run free (without Volta in the loop), described below and in Appendix~\ref{app:formal-methods}:
(i)~support for mathematical constants as immediates, (ii)~support for more instructions (e.g., \texttt{tanh} and \texttt{ldmatrix.trans}), (iii)~asynchronous copies, (iv)~asynchronous Tensor Core instructions and their operand movement
(\texttt{tcgen05}, \texttt{wgmma}, \texttt{ldmatrix}),
(v)~a broader range of barriers,
(vi)~shuffle-based reductions, and
(vii)~packed formats, in which several narrow values share one register.
This support is not free: Volta's trusted computing base grows from 42\,kLoC to 60\,kLoC, and every
newly modeled instruction introduces an additional risk of unsoundness. Appendix~\ref{app:formal-methods} gives details. 

\begin{wraptable}{r}{0.46\textwidth}
\vspace{-\intextsep}
\centering
\small
\setlength{\tabcolsep}{5pt}
\begin{tabular}{@{}lrr@{}}
\toprule
\textbf{Kernel} & \shortstack{\textbf{Provably}\\\textbf{Verified}} & \shortstack{\textbf{Numerically}\\\textbf{Verified}} \\
\midrule
ReLU             & $1.00\times$ & $1.00\times$ \\
GEMV   & $1.15\times$ & $1.15\times$ \\
GEMM & $0.24\times$ & $1.03\times$ \\
FlashAttention & $0.78\times$ & $1.37\times$ \\
Convolution 2D   & $0.77\times$ & $1.35\times$ \\
\bottomrule
\end{tabular}
\caption{Performance of the best provably verified and best numerically verified implementations on an NVIDIA B200 in FP16 using the original Volta.}
\label{tab:verified-vs-unverified-speedups}
\end{wraptable}

With these features, we can verify the unrestricted PTX output of
\aicompilershort{} (without constraining it with Volta in the loop) for every kernel
in the common-kernel suite except the reduction sum and the softmax,
and for six of the ten recently published kernels (FlashAttention, Forgetting
Attention, SageAttention, Lion, and both Mamba-2 primitives). 

Three classes of constructs remain unsupported: atomics, input-dependent control flow, and bit-level value manipulation.
Volta abstracts floating-point operations as real arithmetic, so it cannot
reason about code that computes a function by operating directly on the
bit-level encoding. For example, \aicompilershort{} can implement
floating-point max and min reductions with the integer \texttt{redux.sync}
min/max instructions, after a sign-aware remapping of the bit patterns that
makes integer order agree with floating-point order. The
correctness of this reduction follows from the encoding, which the real-valued
abstraction does not model.
Atomic updates let threads write the same location by design;
checking them requires reasoning about all interleavings of the conflicting
updates, which lies outside Volta's execution model.
Finally, Volta targets structured-CTA kernels without input-dependent control flow, which excludes, for instance, piecewise
implementations that select a different formula depending on the input range.

\section{Conclusion}
\label{sec:conclusion}

We show that large language models can directly lower specialized Triton kernels to competitive PTX that passes randomized differential testing.
Although generation can be orders of magnitude more expensive than millisecond-scale Triton compilation, this cost is incurred once and can be amortized over repeated kernel execution. \Ailowering{} may be particularly compelling for new or lightly supported hardware, where synthesis could be substantially cheaper than developing and maintaining a dedicated optimizing compiler backend. 
More broadly, with sufficiently strong verification, \ailowering{} points toward a new paradigm for generating performance-critical code on hardware accelerators.

\bibliography{iclr2027_conference}
\bibliographystyle{iclr2027_conference}

\appendix
\section{Representative Optimization Examples}
\label{app:optimization-examples}

This section examines six representative kernels for which the generated PTX
outperforms Triton. The first three examples come from the common-kernel suite;
the remaining three come from recently published workloads evaluated on B200.

\subsection{Common Kernels}

\paragraph{Reduction sum: changing the reduction algorithm.}
For a $256\times4096$ input, Triton lowers \texttt{tl.sum} to a conventional
reduction: it accumulates values within threads and warps, combines partial
results through shared memory, and finally uses atomic updates. The generated
PTX instead expresses each row sum as $X\mathbf{1}$ and uses Tensor Cores with
packed FP16 or FP8 inputs and FP32 accumulation. On H100 FP16, two
$16\times8\times16$ \texttt{mma.sync} instructions replace scalar conversion
and reduction chains; three shuffle-and-add stages then reduce the MMA outputs,
with one atomic update per warp and no shared-memory reduction or block barrier.
On L40S FP8, a single warp uses one $16\times8\times32$ MMA and reaches
$1.125\times$ Triton performance. On H100 FP8, the kernel instead scales E4M3
inputs, computes integer-valued partial sums with \texttt{redux.sync.add}, and
combines four warp sums using one barrier, reaching $1.09\times$.

\paragraph{Convolution: reusing overlapping input windows.}
The convolution maps a $32\times64\times56\times56$ input and
$128\times64\times3\times3$ filters to $32\times128\times54\times54$ outputs.
Triton lowers it to a Tensor-Core implicit GEMM with a reduction dimension of
576, but uses scalar loads and shared-memory staging. The generated kernels
instead reuse values shared by neighboring $3\times3$ windows. On H100 FP16,
32-bit loads fetch adjacent pixels, funnel shifts construct neighboring values,
and packed shared-memory stores and double buffering overlap operand preparation
with WGMMA, yielding a $1.90\times$ speedup. On B200 FP8, byte permutations
reuse a small set of 32-bit loads across several filter offsets, while pipelined
input and weight buffers feed \texttt{tcgen05.mma}, yielding $2.23\times$. On
L40S FP8, a circular shared-memory cache provides similar reuse and reaches
$1.17\times$. These kernels retain the same Tensor Core computation as Triton
but eliminate redundant gathers through input reuse, packed accesses, and deeper
pipelining.

\paragraph{H100 FP8 GEMM: reducing WGMMA and epilogue overhead.}
The generated kernel uses larger \texttt{wgmma.mma\_async.m64n256k32}
operations, whereas Triton uses \texttt{m64n64k32}, requiring more WGMMA
instructions to cover the same output tile. Explicit double buffering prepares
the next A/B tile while the current Tensor Core operations execute, and
synchronization occurs only when a shared-memory stage is consumed or reused.
The epilogues also differ: Triton repeatedly stores accumulator fragments to
shared memory, synchronizes, and reloads them to obtain the required layout,
whereas the generated kernel uses a transposing \texttt{stmatrix} variant,
followed by one synchronization and vectorized global stores. These choices
reduce instruction, synchronization, and layout-conversion overhead, explaining
the gain on Hopper and its absence on the other architectures.

\subsection{Recently Published Kernels}

These examples compare each generated implementation with the Triton PTX from
the same run. They show that the benefit comes from jointly reconsidering the
instruction sequence, data movement, synchronization, and decomposition at the
target level, rather than from emitting PTX directly in isolation.

\paragraph{BitDelta: decoding binary weights directly into Tensor Core operands.}
For the evaluated $64\times128$ output tile, Triton unpacks each binary weight,
computes $2b-1$ with integer arithmetic, converts the result to FP16, and stages
the decoded weights in shared memory. The generated PTX constructs packed FP16
representations of $\pm1$ directly: a bit permutation aligns the packed weights
with FP16 sign positions, after which each \texttt{lop3} produces two decoded
values by modifying the sign bits of a packed constant. It also transposes the
internal matrix product so that decoded weights can be written directly to the
tensor-memory operand of \texttt{tcgen05.mma}, eliminating a shared-memory
store-reload cycle. A four-stage activation pipeline, two alternating weight
buffers, and a reduction step increased from 32 to 64 further reduce staging,
synchronization, and loop overhead. The resulting kernel reaches $3.34\times$
Triton performance, and the batched variant reaches $2.51\times$ using the same
strategy.

\paragraph{FlashAttention: aligning softmax ownership with tensor-memory layout.}
Both implementations use Blackwell Tensor Cores for $QK^{\mathsf T}$ and $PV$,
but organize the intervening softmax differently. Triton distributes each score
row across two lanes and therefore requires cross-lane reductions. The generated
PTX assigns each thread a complete 32-score row in the native tensor-memory
layout, making the maximum and sum register-local reductions and eliminating
shuffle instructions. It also combines two 64-row query tiles into one 128-row
tile, so each key/value tile serves twice as many queries. Packed \texttt{f32x2}
arithmetic reduces scalar work, while double buffering and overlap between
normalization updates and matrix operations reduce block barriers from three to
one per loop iteration. Together, these changes yield a $1.37\times$ speedup.

\paragraph{FlashSinkhorn: partitioning the reduction across independent warps.}
For $n=m=1024$ and coordinate dimension 64, Triton performs 32 sequential
log-sum-exp updates, with eight warps cooperating on each $32\times32$ tile.
The generated PTX assigns each warp an independent 128-point slice and keeps
separate maximum and sum accumulators. Each warp executes eight iterations
without block barriers or shuffles in the loop, and the partial results are
combined only once at the end. The fixed coordinate tile remains in registers,
and the original FP16 coordinates feed \texttt{mma.sync.m16n8k16} directly,
replacing Triton's FP16-to-FP32 conversions and TF32 \texttt{m16n8k8}
operations. Doubling the reduction extent per MMA also halves the number of
matrix instructions per dot product. The resulting kernel passes the evaluation
tolerances and reaches $1.55\times$ Triton performance.

\section{Verification and Volta Extensions}
\label{app:formal-methods}

The set of features supported by Volta is not sufficient for verifying the correctness of PTX programs optimized for recent GPU architectures (see Table~\ref{tab:verified-vs-unverified-speedups}).
As a proof of concept, we extend Volta to support the most common instructions emitted by AI models.
We do not claim our extensions to be sound or complete.
We describe the extensions below, but their formalization is left as future work.

The main addition is support for asynchronous operations.
Instructions such as asynchronous copies (\texttt{cp.async}) introduce a new class of memory hazard, beyond cross-thread data races on synchronous memory accesses.
Volta tracks data races at byte granularity by tracking the accesses of reader and writer threads between synchronization barriers.
We handle asynchronous data movement by also tracking asynchronous operations at both the source and the destination.
An asynchronous operation is considered in flight from its initiating instruction until the execution of an appropriate barrier instruction.
Our extended Volta reports a memory hazard if an asynchronous data movement overlaps with either a synchronous or an asynchronous operation.

The next significant addition is support for the long tail of instructions commonly found in generated kernels.
Of those, Tensor Core operations, \emph{e.g.}, the \texttt{tcgen05}, \texttt{wgmma}, and \texttt{ldmatrix} instruction families, are the most complex due to their large set of configurations and their interaction with asynchronous operations.
We further implement missing math operators, such as \texttt{tanh} and \texttt{exp2}, which we express in terms of exponentials to leverage the existing solver simplification rules.
We add support for packed representations, such as 16x2 and 8x4 packed values, as well as arithmetic, conversion, and load/store operations on them.

Finally, we found it necessary to explicitly support common patterns.
For instance, in shuffle reductions, all threads may concurrently write the \emph{same} value to the same memory address. This well-defined data race is common in kernels and semantically harmless, since every thread stores the same value.
To handle that, we modify Volta to establish the equality of the values being written at the time it checks for the race, invoking the equivalence checker in the middle of the symbolic execution.

Our extensions are sufficient to generate kernels that match or exceed the Triton compiler, but there remain significant intrinsic limitations.
First, Volta does not support data-dependent control flow.
Although rare, data-dependent control flow can be found in kernels for patterns such as segmented polynomial approximations or conditional updates during streaming computations.
Atomic operations are also out of scope to avoid having to consider all possible thread interleavings.
Finally, due to the infinite-precision floating-point representation, kernels that rely on properties of the bit layout of floating-point numbers might be rejected by Volta despite producing the expected result with finite-precision arithmetic.



\end{document}